\documentclass[conference]{IEEEtran}
\IEEEoverridecommandlockouts
\usepackage{cite}
\usepackage{bm} 
\usepackage{bbm}
\usepackage{amsmath,amssymb,amsfonts}
\usepackage{algorithmic}
\usepackage{graphicx}
\usepackage{textcomp}
\usepackage{xcolor}
\usepackage{caption}
\usepackage{subfigure}
\def\BibTeX{{\rm B\kern-.05em{\sc i\kern-.025em b}\kern-.08em
    T\kern-.1667em\lower.7ex\hbox{E}\kern-.125emX}}
\begin{document}

\title{A Graph Regularized Point Process Model For Event Propagation Sequence
\thanks{The 4th to 7th authors are corresponding authors.}}

\author{\IEEEauthorblockN{1\textsuperscript{st} Siqiao Xue}
\IEEEauthorblockA{\textit{Intelligent Engine Technologies} \\
\textit{Ant Group}\\
Hangzhou, China \\
siqiao.xsq@alibaba-inc.com}
\and
\IEEEauthorblockN{2\textsuperscript{nd} Xiaoming  Shi}
\IEEEauthorblockA{\textit{Intelligent Engine Technologies} \\
\textit{Ant Group}\\
Hangzhou, China \\
peter.sxm@alibaba-inc.com}
\and
\IEEEauthorblockN{3\textsuperscript{rd} Hongyan Hao}
\IEEEauthorblockA{\textit{Intelligent Engine Technologies} \\
\textit{Ant Group}\\
Hangzhou, China \\
hongyanhao.hhy@alibaba-inc.com}
\and
\IEEEauthorblockN{4\textsuperscript{th} Lintao Ma}
\IEEEauthorblockA{\textit{Intelligent Engine Technologies} \\
\textit{Ant Group}\\
Hangzhou, China \\
lintao.mlt@antgroup.com}
\and
\IEEEauthorblockN{5\textsuperscript{th} Shiyu Wang}
\IEEEauthorblockA{\textit{Digital Finance Technologies} \\
\textit{Ant Group}\\
Hangzhou, China \\
weiming.wsy@antgroup.com}
\and
\IEEEauthorblockN{6\textsuperscript{th} Shijun Wang}
\IEEEauthorblockA{\textit{Intelligent Engine Technologies} \\
\textit{Ant Group}\\
Seattle, US \\
shijun.wang@antgroup.com}
\and
\IEEEauthorblockN{7\textsuperscript{th} James Zhang}
\IEEEauthorblockA{\textit{Intelligent Engine Technologies} \\
\textit{Ant Group}\\
New York, US \\
james.z@antgroup.com}
}
\maketitle

\begin{abstract}
Point process is the dominant paradigm for modeling event sequences occurring at irregular intervals. In this paper we aim at modeling latent dynamics of event propagation in graph, where the event sequence propagates in a directed weighted graph whose nodes represent event marks (e.g., event types). Most existing works have only considered encoding sequential event history into event representation and ignored the information from the latent graph structure. Besides they also suffer from poor model explainability, i.e., failing to uncover causal influence across a wide variety of nodes. To address these problems, we propose a Graph Regularized Point Process (GRPP) that can be decomposed into: 1) a graph propagation model that characterizes the event interactions across nodes with neighbors and inductively learns node representations; 2) a temporal attentive intensity model, whose excitation and time decay factors of past events on the current event are constructed via the contextualization of the node embedding. Moreover, by applying a graph regularization method, GRPP provides model interpretability by uncovering influence strengths between nodes. Numerical experiments on various datasets show that GRPP outperforms existing models on both the propagation time and node prediction by notable margins.
\end{abstract}


\section{Introduction}
Multitudes of irregular event sequences, labeled with event timestamps and types, are being generated in digital world at every moment. For example, activities of transactions in financial markets, purchases in e-commerce platforms, visits to hospitals and postings in social medias can all be formulated as discrete event sequences happening at irregular intervals. It is essential to model these complex event sequence dynamics so that 
accurate prediction or intervention can be carried out subsequently depending on the context.

Point process \cite{daley2003}, characterized by the intensity function that measures the instantaneous probability of occurrence, has long been the standard tool in modeling event sequences. Poisson process  \cite{palm1943inten}, a simple type of point process, has a constant intensity function and Hawkes process \cite{hawkes1971spectra} constructs the conditional intensity function to recognize self-excitation 
between events as an enhancement to Poisson process. Recently, RNN-based models, e.g.,Recurrent Marked Temporal Point Process (RMTPP) \cite{du2016recurrent} and Neural Hawkes Process (NHP) \cite{mei2017neural}, are proposed to model the dynamics of event sequences, which achieve significant progresses in learning and prediction tasks. Besides the variants of attention-based Hawkes process \cite{sapp2020,thp2020} use deep attention structure to capture the long-term dependencies of the sequences.

\begin{figure*}
\centering
\includegraphics[width=14cm]{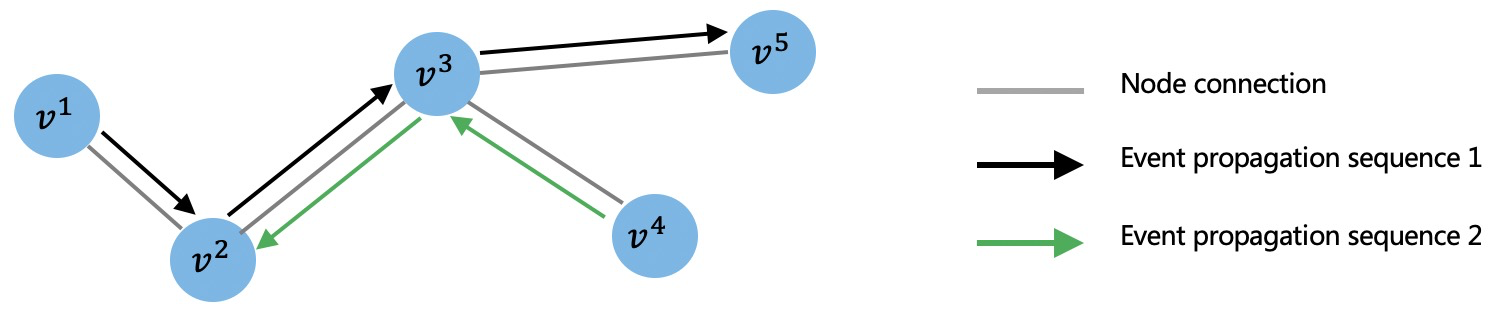}
\caption{\footnotesize An example of two event propagation sequences: $(v^1, v^2, v^3, v^5)$ and $(v^4, v^3, v^2)$ in the graph, where each node corresponds to a specific event type and each edge represents influences between the two event types. $v^i$ denotes the event propagating to the $i$-th node in the graph.}
\end{figure*}

In this paper, we focus on the problem of modeling the dynamics of graph event sequence, where the event propagates in a directed weighted graph and \textbf{the mark is denoted as the node in the graph}. The examples of such sequences include meme (news articles and blog posts) cascade sequences over public media, where the websites correspond to the nodes in the graph network. As denoted in \cite{dyrep2019,gbtpp2020}, abstracting the information from the \textbf{mark} (i.e., event type) space helps discover the structure in the point process effectively and efficiently. Sharing the same motivation, we build a graph propagation process that encodes evolving structural properties of the observed sequence into the node embedding of the graph, and construct a latent intensity model, which uses the learnt node embedding to capture both short-term and long-term complex dependencies of the sequences. Besides, 
motivated by the work in \cite{zhou2013} that incorporates social structure as constraints to learn influence patterns between users in social networks, we integrate the prior structural knowledge into the modeling process to learn a more reasonable and interpretable influence pattern between nodes.

We propose a graph regularized point process (GRPP) to model event propagation sequences. Compared with existing works, we make two main contributions:
\begin{itemize}
    \item To capture the mutual effects of the changing graph structure and temporal interactions between nodes, we employ two sub-modules to model separately the structural and temporal components of the event propagation process in graph: 1) a graph propagation model that characterizes the event interactions across nodes with neighbors and inductively learns node representations; 2) a temporal attentive intensity model, whose excitation and time decay factors of the past events imposing on the current event are constructed via contextualization of the node embedding, which greatly facilitates the learning of dependencies from the distant past. 
    \item By leveraging a regularization method, which imposes a prior connection graph to guide the learning of the mutually-exciting coefficients, a.k.a, infectivity matrix, GRPP is able to maintain the model interpretability \cite{rppn2019} by uncovering causal influence between nodes. The uncovered strength of influence is critical as it provides the understanding of how the model works internally and facilitates decision making.
\end{itemize} 
The proposed GRPP model demonstrates consistent improvement over state-of-the-art baselines on both synthetic and real datasets on both the learning and prediction tasks. The code will be released on Github upon the internal approval of the company.

\section{Background and Preliminaries}
\subsection{Point Process}
A point process \cite{daley2003} is a stochastic process whose realization consists of a list of discrete events localized in times $\{t_i\}_{i \in \mathbb{N}^*}$ where $\forall i \in \mathbb{N}^*, t_i < t_{i+1}$. A point process can be best characterized by a conditional intensity function $\lambda_t :=  \lim_{\Delta \rightarrow 0} \frac{\mathbb{E}\left[ N_{t+\Delta} - N_t \vert \mathcal{H}_t   \right]}{\Delta}$
, where $\mathbb{E}\left[ N_{t+\Delta} - N_t \vert \mathcal{H}_t   \right]$ represents the expected number of events that occur in the interval $(t, t + \Delta]$ given history $\mathcal{H}_t$. 

The univariate Hawkes process has a conditional intensity $\lambda_t = \mu + \sum_{t_i<t} g(t-t_i)$
, where $\mu$ is the base intensity, $t_i$ is the $i$-th event’s occurrence time and $g(\cdot)$ is a kernel function measuring the influence of the previous events on the current event. The univariate Hawkes process can be extended to multi-dimensional Hawkes Process (MHP) to handle multiple types of events happening sequentially. Specifically, for a $K$-dimensional event sequence, the conditional intensity function of the $i$-th dimension is 
\begin{align}
    \lambda_i(t) = \mu_i + \sum_{j=1}^K \alpha_{ij} \sum_{t_j<t}g(t-t_j) ,
    \label{eqn:hawkes}
\end{align}
where $\mu_i$ is the base intensity of the dimension $i$, $\alpha_{ij} \in \mathbf{A}$ measures the \emph{causal influence} of dimension $j$ to dimension $i$, and $\mathbf{A} \in \mathbb{R}^{K \times K}$ is called the \emph{infectivity matrix} across event dimensions \cite{rppn2019}.


\subsection{Difference and Connection to Existing Works}
From the perspective of the methodology, we consider existing works \cite{graphreg2018,gbtpp2020,dyrep2019,thp2020} to be related to our work as they leverage the information of graph structures. However, there are both differences and similarities between our work and the existing works. 

Specifically, \cite{graphreg2018,thp2020} propose a spatial-temporal point process by encoding spatial graph information from the extra mark of the event tuple which denotes geographic locations, i.e., $(t_i,v_i, \xi_i)$, where $t_i, v_i, \xi_i$ signify the event timestamp, the event type and the geographic information, respectively.  Meanwhile, the event description in our problem is confined to only the timestamp and type, i.e., $(t_i,v_i)$ without introducing any extra mark, while the graph we build has nodes that correspond to event types instead of geographic locations. While the recent work \cite{gbtpp2020} proposes a Graph Biased Temporal Point Process (GBTPP) which employs the combination of RNN and graph embedding, there are essential differences between GBTPP and our work. Specifically, we propose a graph model with second-order proximity preserved, via a graph attention mechanism, to incorporate the influence between nodes and the node embedding is dynamically updated while GBTPP learns a graph only with first-order proximity and the node embedding is statically learnt in a separate stage, where the information of the dynamics between the events in different nodes is lost. We also utilize the latent attentive intensity model to incorporate historical influence of events while GBTPP constructs the intensity function without considering the complex dependencies.

\begin{table*}[h!]
\centering
\caption{Comparison of our model with state-of-the-art approaches}
 \begin{tabular}{|c  c c c|} 
 \hline
 Key Properties & Our Model & GBTPP & RMTPP \\ 
 &   &   &   /NHP  \\[0.5ex]
 \hline\hline
  Models Graph Propagation & $\checkmark$ & $\checkmark$ & \texttimes \\ 
 \hline
  Graph Information &  2nd-order Proximity  &  1st-order Proximity & \texttimes \\
 \hline
 Learns Node Representation & Dynamic  &  Static & \texttimes \\
 \hline
  Attention Mechanism  & Attentive Latent Intensity  &  \texttimes & \texttimes \\
 \hline
   Graph Recovery & $\checkmark$  & \texttimes & \texttimes \\
  (Causal Matrix Recovery) & &  &  \\
  \hline
  Predicts Times
 & $\checkmark$ & $\checkmark$ & $\checkmark$ \\
 \hline
  Predicts Nodes (Types)
 & $\checkmark$ & $\checkmark$ & $\checkmark$ \\ 
 \hline
\end{tabular}
\label{table:compare}
Å\end{table*}

Furthermore, different from the concurrent work \cite{thp2020,sapp2020}, which applies the transformer structure to model the point process, the intensity mechanism in our model evolves in a \emph{latent space} instead of the actual event space whose dimension equals to the number of event types, which is often hard to process under large or sparse feature space. Lastly, \cite{dyrep2019} focuses on graph representation learning over temporal dynamic graph as a latent mediation process while our model targets at learning a point process with interpretability with strong prediction accuracy by utilizing the graph information. 

Table \ref{table:compare} provides qualitative comparison between the most related state-of-the-art models and our framework.

\subsection{Notations and Problem Formulation}
We assume a system with $K$ event types, denoted as $\mathcal{K}=\{1,2,...,K\}$ and there exists a latent \emph{directed} graph $\mathcal{G}=(\mathcal{V}, \mathcal{E})$ with node set $\mathcal{V}=\mathcal{K}$ and edge set $\mathcal{E}=\{e_{ij}\}_{K \times K}$. The nodes in $\mathcal{G}$ correspond to the dimensions of the event space and the edges signify the strength of the endogenous influence across different nodes. Specifically, edge $e_{uv}$ indicates that the node $u$ is the parent of the node $v$ (i.e., an event with type $u$ could cause an event with type $v$). Following the previous work \cite{latent2019}, we only consider the first parent as the true parent. Suppose we are given an observed event propagation sequence $\mathcal{O}_{t}=\{(t_i, v_i): t_i <t\}_{i=1}^N$ of $N$ events up to time $t$, where each event propagates to node $v_i \in \mathcal{K}$ at time $t_i$. For the convenience of notation, we use $k$ to denote the index of nodes in graph $\mathcal{G}$, or equivalently the index of event types in $\mathcal{K}$. 

Given the observed event sequence $\mathcal{O}_{t}$, we aim to recover the latent graph 
$\mathcal{G}$, and learn a predictor that predicts next event's propagation time and node by leveraging the structural information of $\mathcal{G}$.

\section{Proposed Model: GRPP}
The main idea of GRPP is to build a unified architecture that ingests evolving information over graphs to uncover the latent structure of the point process. Therefore, inspired by the work in \cite{dyrep2019}, we design a point process model which is parameterized with inductive node representations. By leveraging the node representations, we construct an attentive intensity model that captures both the temporal dynamics and structural dependencies in the event sequence. Specifically, GRPP consists of a positional encoder for producing the order-dependent embedding for the event sequence and two specially designed modules: 
\begin{itemize}
    \item A graph propagation 
model for parameterizing the event propagation across nodes and learning node representations.
\item An attentive intensity model for capturing the long-term and complex dependencies intrinsic to the event propagation sequence. 
\end{itemize}
\begin{figure*}[!htp]
    \centering
    \includegraphics[width=0.65\textwidth]{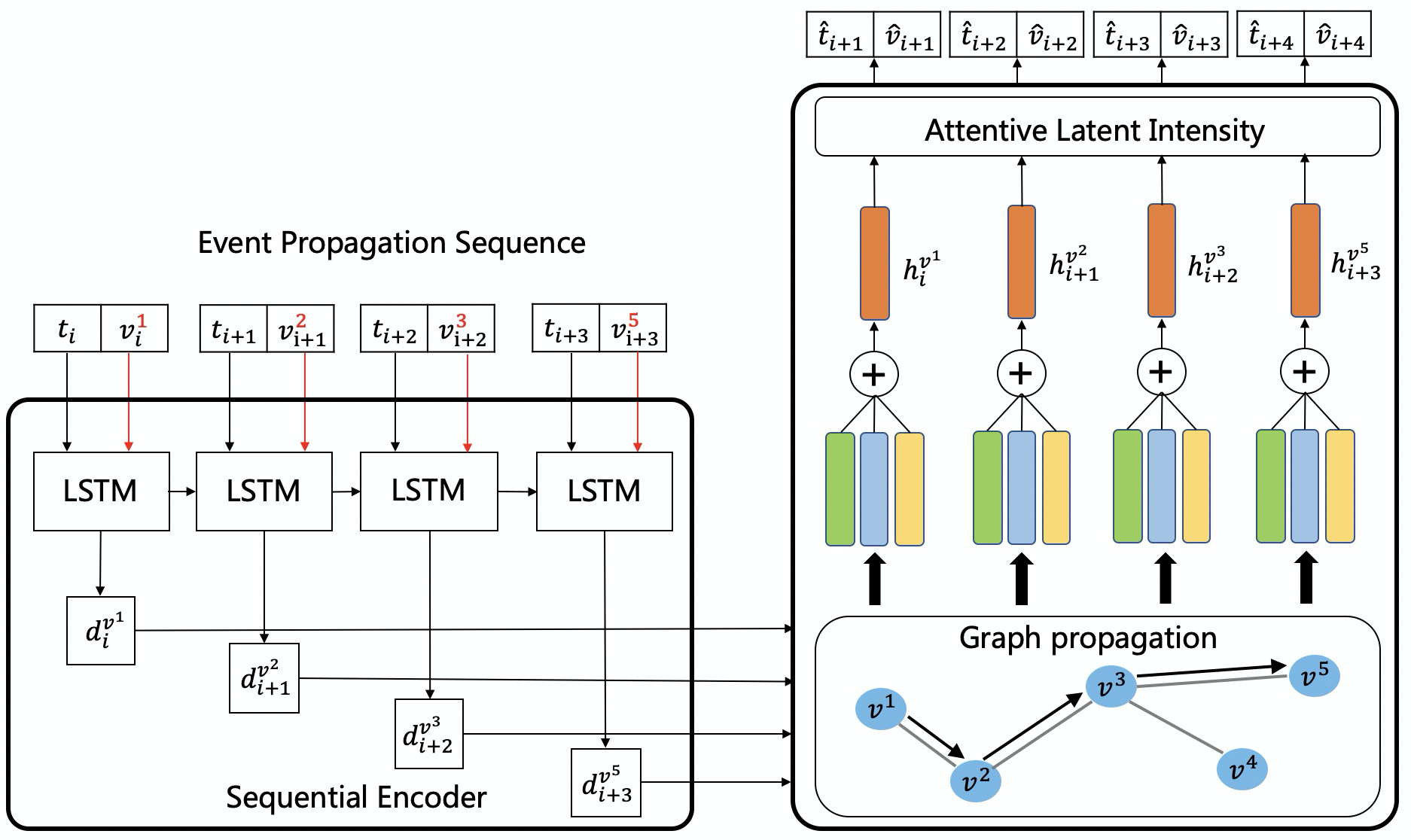}
    \caption{\footnotesize Framework of GRPP: Assume an observed sequence propagating through nodes $v^1, v^2, v^3, v^5$ sequentially. They are fed through an LSTM layer to obtain sequential encodings, and then fed into a graph propagation module characterizing the event interactions across nodes with neighbors, while inductively learning node representations. In the end, an attentive module captures the intensity dynamics in the latent space. The output is the prediction of the next propagation time and node. For the convenience of notation, we use $v^i$ to denote the event propagating to the $i$-th node in the graph.}
    \label{fig:my_label}
\end{figure*}

\subsection{Model Architecture}

\subsubsection{Sequential Encoder}
For an event propagating from node $u$ to $v$ at $t_i$, we first use a linear embedding layer to obtain a unique dense embedding for event type $\bm{x}^{v}_{t_i} \in \mathbb{R}^m$, and then pass it to a positional encoding layer composed of LSTM cells, to capture the order-dependent information of the event sequences. Specifically, the encoding layer maps the event type embedding $\bm{x}_{t_i} \in \mathbb{R}^m$ to a positional encoding vector $\bm{d}^v_{t_i} \in \mathbb{R}^d$ by $\bm{d}^{v}_{t_i}=\textbf{LSTM}([\bm{d}^{u}_{t_{i-1}};\bm{x}^v_{t_i}])$
We prefer LSTM to other structure because the cell state sums activities over time, which overcomes the problem of diminishing gradients and at the same time better captures long-term dependencies of the event sequences than vanilla RNNs.

\subsubsection{Graph Propagation}
The rationale of this sub-model is that the observed propagation sequences are the realizations of a latent point process that governs the changes in the topological structure of graph and interactions between the nodes in the graph. After the occurrence of an event, we update its corresponding node representation to capture the effect of propagation based on the principles of localized embedding propagation as well as self-propagation \cite{dyrep2019}.

Given the sequential embedding $\bm{d}^v_{t_i}$, we evolve its node representation vector $\bm{h}^v \in \mathbb{R}^d$ as:
\begin{align}
    \bm{h}^{v}_{t_i}=  \underbrace{\bm{W}_1 \bm{s}^{u}_{t_{i-1}}}_{\small\text{Local propagation}} + \underbrace{\bm{W}_2 \bm{h}^{v}_{t_{i-1}}}_{\small\text{Self-propagation}} +  \underbrace{\bm{W}_3 \bm{d}^{v}_{t_i}}_{\small \text{Exogenous drive}}
    \label{eqn:node}
\end{align}
\noindent$\diamond$ \quad $\bm{W}_1 \bm{s}^{u}_{t_{i-1}}$ corresponds to the localized propagation process and $\bm{s}^{u}_{t_{i-1}} =\sum_{z \in \mathcal{N}_u} q_{z} \bm{h}^z_{t_{i-1}}$ is obtained from an aggregation on node $u$'s neighborhood $\mathcal{N}_u$ with a graph attention method proposed in \cite{gat2018,dyrep2019}, where $q_z=\frac{\exp(score(\bm{h}^u_{t_{i-1}},\bm{h}^z_{t_{i-1}})}{\sum_{p \in \mathcal{N}_u}\exp(score(\bm{h}^u_{t_{i-1}}\text{,} \bm{h}^p_{t_{i-1}}))}$ signifies the attention weight for the neighbour $z$ during the propagation and \emph{score} is the attention score method \cite{luong2015}. The localized propagation process, where the information propagates from $u$'s neighbours to $v$, captures the influence from the nodes at \emph{second-order proximity} passing through the other node involved in the event.

\noindent$\diamond$ \quad $\bm{W}_2 \bm{h}^{v}_{t_{i-1}}$ refers to a self-propagation that updates its own previous node embedding. 

\noindent$\diamond$ \quad $\bm{W}_3 \bm{d}^{v}_{t_i}$ serves as an exogenous influence from the current event.

With our new setup, the representation of the involved node is updated to capture the rich structural properties of the event sequence. We then define a latent intensity function characterized by the temporally evolving node  representations as follows.

\subsubsection{Latent Intensity}
Different from the concurrent work \cite{thp2020,sapp2020}, we model the intensity dynamics in a latent space $\mathbb{R}^d$. The \emph{latent conditional intensity} $\lambda^h$ evolves as Hawkes process, except that the base intensity, the activation and decay factors are characterized using neural networks and are learnt from data. For simplicity, in this section, we use $h_i$ to denote the node embedding vector updated in Equation \ref{eqn:node} and ignore the notation of its corresponding node. 

For $t \in [t_i, t_{i+1})$, we define the \emph{latent conditional intensity} vector $\bm{\lambda}^h_{t} \in \mathbb{R}^d$ as
\begin{align}
    \bm{\lambda}^h_{t} = \sum_{j=0}^{i-1} \bm{\alpha}_{j,i} \odot \bm{\delta}_{j,i} (t - t_j) + \bm{\mu}_{t_i}, \quad i \ge 1
    \label{eqn:implied_int}
\end{align}
\noindent$\diamond$ \quad $\bm{\mu}_{t_i}= \sigma(\bm{W}^{\mu} \bm{h}_{t_i} + \bm{b}^{\mu}) $ is the base intensity vector, representing the probability of occurrence of event propagation without considering the history information.

\noindent$\diamond$ \quad $\bm{\alpha}_{j,i} =\beta_{j,i} \bm{h}_{t_j}$ represents the initial excitation of the $j$-th 
    event on the $i$-th event. It is computed as the weighted hidden representation with $\beta_{j,i}$, which is the attention weight measuring the relevance of corresponding node representation of $j$-th event and the $i$-th event, calculated using a softmax transformation of $\omega_{j,i}$ in the form of a score:
    \begin{equation}
        \omega_{j,i} = \bm{V} \tanh(\bm{W}^{\omega}[\bm{h}_{t_i}; \bm{h}_{t_j}]), 
        \beta_{j,i} = \frac{\omega_{j,i}}{\sum_{u=0}^{i-1} \omega_{u,i}}  
        \label{eqn:beta}
    \end{equation}
Note that $\bm{\alpha}_{j,i}$ aims to capture the \emph{long-term dependencies of excitation effects} over the propagation sequence. Besides, during the learning process, $\bm{\alpha}_{j,i}$ could become negative to capture the inhibition effect \cite{mei2017neural} because $\beta_{j,i} \in (0, 1)$ and $\bm{h}_{t_i} \in (-1, 1)$.

\noindent $\diamond$ \quad $\bm{\delta}_{j,i} =\sigma\left(\bm{W}^{\delta}([\bm{h}_{t_j}; \bm{h}_{t_i}]))+\bm{b}^{\delta}\right)$ represents the time decay factor. Specifically, we directly encode the $j$-th event's and the $i$-th event's node representations, multiplied by the exponential decay, to measure the relative time decay influence of the $j$-th event on the $i$-th event. 

Finally, we apply a Feed-Forward Network (FFN) with \textbf{softplus} activation to map the \emph{latent intensity} $\bm{\lambda}^h_{t_i} \in \mathbb{R}^d$ to the event intensity $\bm{\lambda}_{t} \in \mathbb{R}^K$.

\subsubsection{Graph Regularization}
\label{section:reg}
Using $\mathbf{H} \in \mathbb{R}^{K \times d}$ to denote the node embedding matrix and $\mathbf{\Omega} \in \mathbb{R}^{d \times d}$ a matrix to be learned. Given the learned node embedding, we compute the adjacent matrix $\mathbf{A}$ of $\mathcal{G}$ as $\mathbf{A} = \mathbf{H} \mathbf{\Omega}\mathbf{H}^T$. By definition, the edges of $\mathcal{G}$ describe the influence between nodes, and therefore the adjacent matrix 
$\mathbf{A}$ coincides with the infectivity matrix defined in Equation \ref{eqn:hawkes}.

Adopting the learning framework proposed in spatial-temporal point process \cite{graphreg2018,thp2020}, we introduce a connection matrix $\mathbf{E}=\{ e^{'}_{ij} \} \in \mathbb{R}^{K \times K}$ as a prior to represent the local causal relations which \emph{exists pervasively and agrees with human intuition} across different event dimensions. For example, an intense sports game is more likely to cause a rest than going to the movies. 
As the common practice in previous works \cite{gbtpp2020,rppn2019}, given the observed sequence, we use the empirical estimation for the entries in $\mathbf{E}$,  i.e., $e_{ij}=\frac{N_{ij}}{N_{max}}$ where $N_{ij}$ is the number of events propagated from node $i$ to $j$ and $N_{max}$ is the total number of event propagated between the two node.
Finally we adopt the KL-divergence as the distance measure to minimize the difference between the $\mathbf{E}$ and $\mathbf{A}$ so that the connection matrix is used as a constraint to guide the learning of infectivity matrix.

\subsection{Training and Inference}

We integrate the prior structural knowledge into modeling the event sequence by imposing the connection matrix $\mathbf{E}$ as constraints in learning infectivity matrix $\mathbf{A}$, with which we formulate the loss function as follows:
\begin{align}
    \min - \mathcal{L}_{nll} + \gamma \mathcal{L}_{graph}
    \label{eqn:loss}
\end{align}, 
where $\mathcal{L}_{nll}=- \sum_{t_i < T} \log \lambda^{u_i}_{t_i} +  \int_0^T \lambda_t dt$ is the negative log-likelihood \cite{daley2003,rasmussen2013} for event sequences, $\gamma$ is a hyper-parameter and $\mathcal{L}_{graph}= \Vert \mathbf{A} - \mathbf{E} \Vert_{\mathbf{KL}} $ is a regularization term clarified previously in the beginning of Section \ref{section:reg}. Model parameters are learned by optimizing Equation~\eqref{eqn:loss} with the backpropagation through time (BPTT) \cite{gbtpp2020} technique, which maintains the dependencies between events while avoiding gradient-related problems. The density of the next propagation time is $p_{i+1}(t | \mathcal{H}_{i+1})= \lambda(t) \exp{\left( -\int_{t_{i}}^t \lambda(s) ds \right)}$.
 The predictions of the next propagation's time and node are obtained by taking the expectation, i.e., $\hat{t}_{i+1}=\int_{t_{i}}^\infty t p_{i+1}(t) dt, \hat{k}_{i+1} = \arg\max_{k} \int_{t_{i}}^\infty \frac{\lambda_k(t)}{\lambda(t)} p_{i+1}(t) dt$.
 

\section{Experiments}
\subsection{Experimental Setup}
\subsubsection{Datasets and Metrics}
We conduct experiments using two synthetic datasets and two public datasets, i.e., Higgs\cite{gbtpp2020} and Meme \cite{mem2014}, to evaluate the predictive performance of the proposed model. We use accuracy ratio and RMSE as the metrics for the propagation node and time prediction tasks. To clarify again, the marks correspond to the nodes in the graph.
\begin{itemize}
    \item \textbf{Synthetic datasets}
The synthetic data of graph event propagation process are generated from the multivariate Hawkes process (MHP), which has been widely used to model the generative process of user behavior of social networks \cite{farajtabar2015coevolve,rppn2019}. In our setting, each dimension of Hawkes process corresponds to an individual node and the causal influence between nodes are explicitly modeled. Following the same setup as \cite{gbtpp2020}, we simulate a 10-dimensional MHP and a 100-dimensional MHP (\textbf{syn-10d} and \textbf{syn-100d}). The base intensity is sampled from a uniform distribution between $[0, 0.001]$ and the infectivity matrix is generated from $\mathbf{A}=\mathbf{UV}^T$, where $U$ and $V$ are $10 \times 1$ (and $100 \times 9$) matrix with entries between $[(i-1)+1: (i+1), i]$ (and $[10(i-1)+1: 10(i+1), i]$ if $U$ and $V$ are $100 \times 9$), $i=1,...,9$ sampled uniformly between $[0, 0.1]$ and all the other entries are assigned zero. 

\item \textbf{Higgs}
The Higgs dataset has been built after monitoring the spreading processes on Twitter before, during and after the announcement of the discovery of a new particle with the features of the elusive Higgs boson on 4th July 2012. As the common practice in previous work \cite{gbtpp2020}, we study the retweet propagation process by using the largest strongly connected component which contains 984 nodes (users) and 3,850 edges in the retweet network and train the model on the retweet activities.

\item \textbf{Meme}
The dataset tracks meme diffusion process over public media, containing millions of news articles and blog posts. Following the same procedure as \cite{gbtpp2020}, we extract top 500 popular sites and 62,000 meme diffusion cascades among them. We also study the meme propagation process between websites.

\end{itemize}

\subsubsection{Baselines and Training Details}
We herein evaluate the predictive performances of our model against
\begin{itemize}
    \item Multivariate Hawkes Process(\textbf{MHP}) \cite{hawkes1971spectra}: a self-exciting multivariate point process, which assumes that past events can additively increase the probability of subsequent events.
    \item Recurrent Marked Temporal Point Process (\textbf{RMTPP})\cite{du2016recurrent}: a RNN-based model to learn a representation of influences from past events.
    \item Neural Hawkes Process (\textbf{NHP})\cite{mei2017neural}: a continous-LSTM model to capture the dependencies of events.
    \item Graph Biased Temporal Point Process (\textbf{GBTPP}) \cite{gbtpp2020}: the most related model which leverages the structural information from graph representation learning to model graph propagation sequences.
\end{itemize}


Each dataset is splited into train set, validation set and test set with a ratio of 3:1:1.  
We conduct all the experiments with a hyper-parameter grid search strategy and choose the hyperparameters that achieve the best results on the validation dataset. We use the Adam optimizer to train the model. The batch size is set to 256, and the learning rate is set as 0.001. The dropout rate in LSTM is set to 0.2. The dimension of hidden state of all LSTM units $d$ was set as 128, and the embedding dimension of the event $m$ is set as 128 as well. The regularization parameter $\mu_l=0.01$.


\subsection{Experiment Results}
\begin{figure*}[htp]
\centering
\subfigure[Propagation node prediction.]{
    \includegraphics[width=7.0cm]{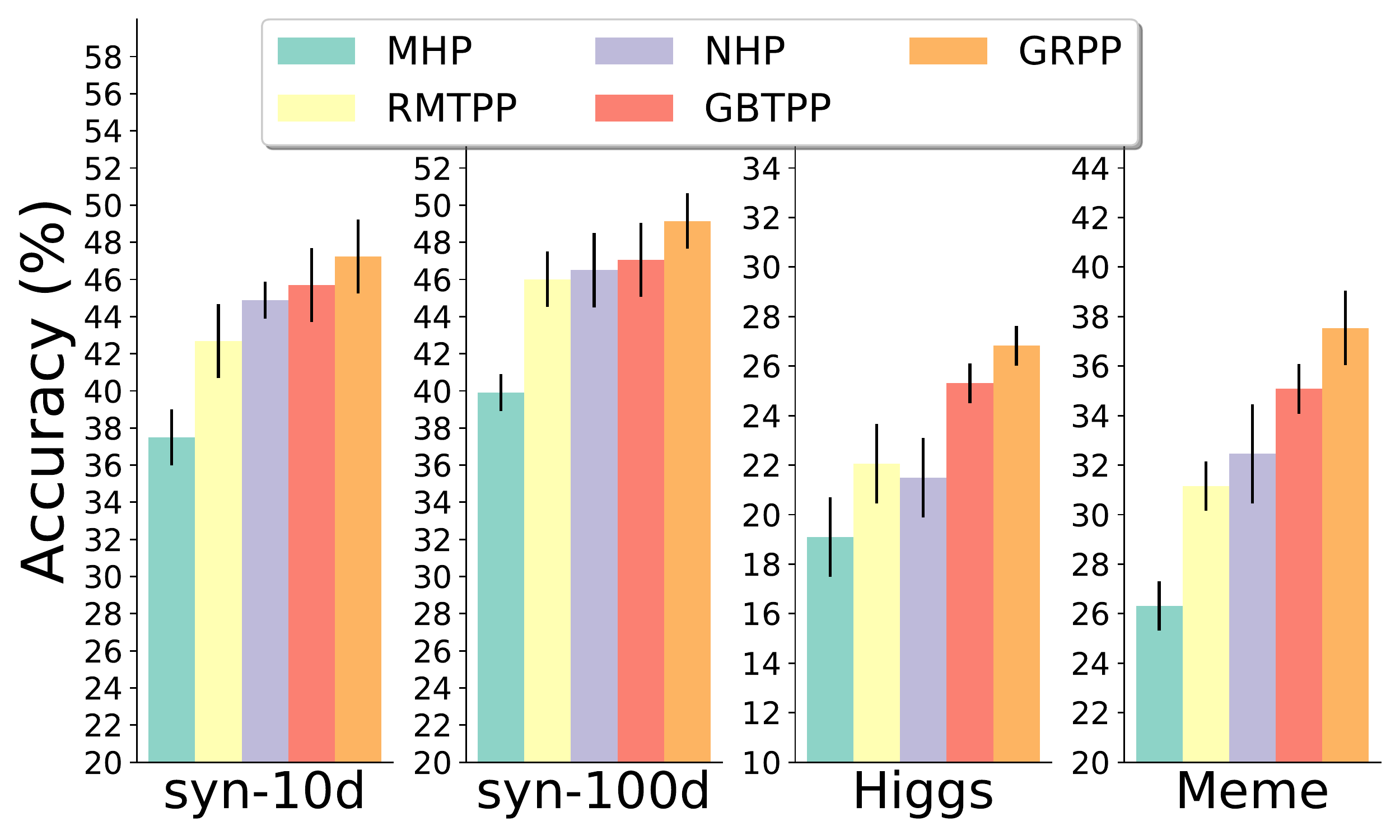}}
\quad
\subfigure[Propagation time prediction.]{
    \includegraphics[width=7.0cm]{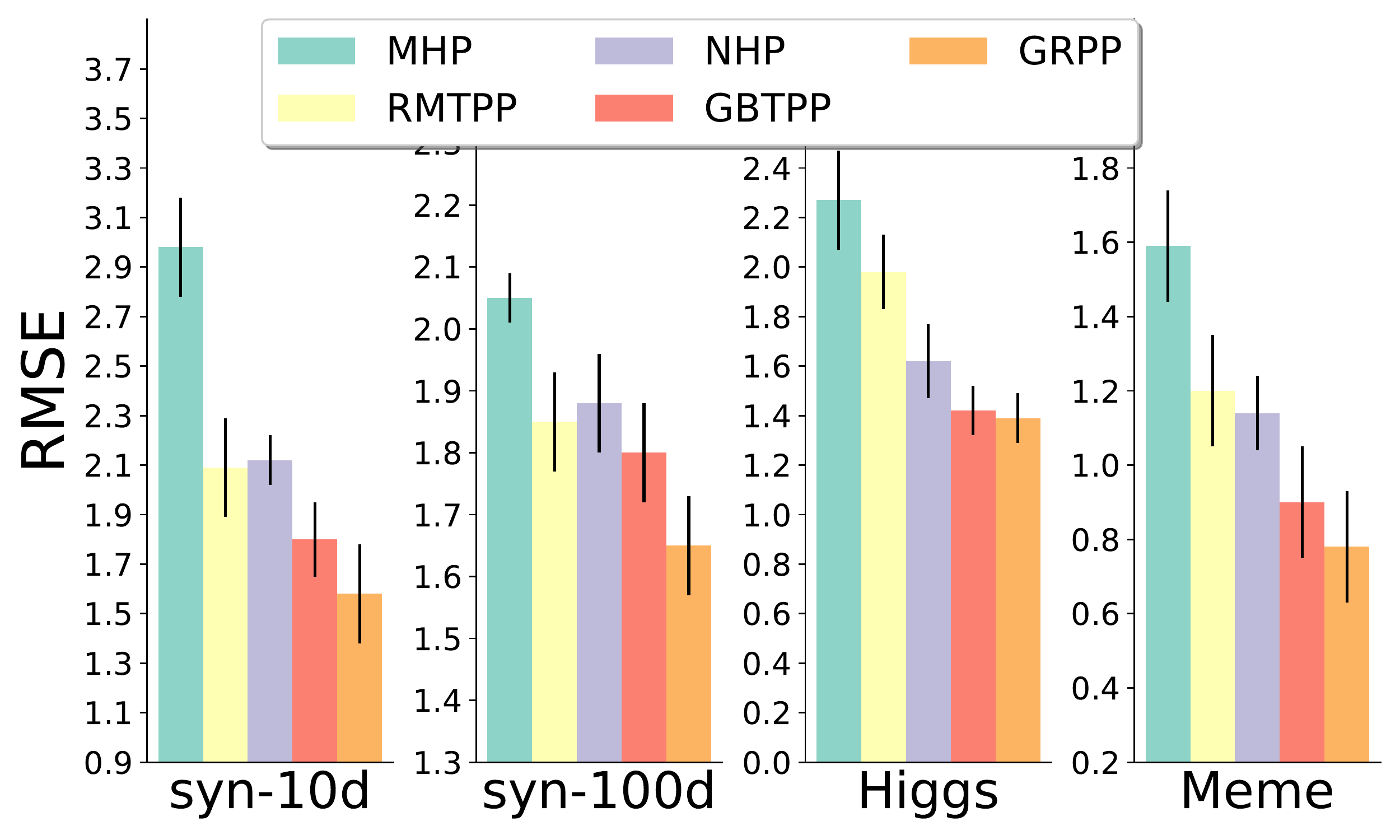}}
\caption{Propagation prediction performances on synthetic and real datasets. Based on a same train-valid-test splitting ratio, each dataset is sampled five times to produce different train, validation and test sets. Error bars are generated according to these experiments.}
\label{fig:res}
\end{figure*}

\begin{figure*}[htp]
\centering
\subfigure[Comparison of infectivity matrix on syn-10d dataset: left(ground-truth) vs right(recovered from GRPP). Entry$(i,j)$ in the matrix denotes the causal influence from node $i$ to node $j$.]{
    \includegraphics[width=10cm]{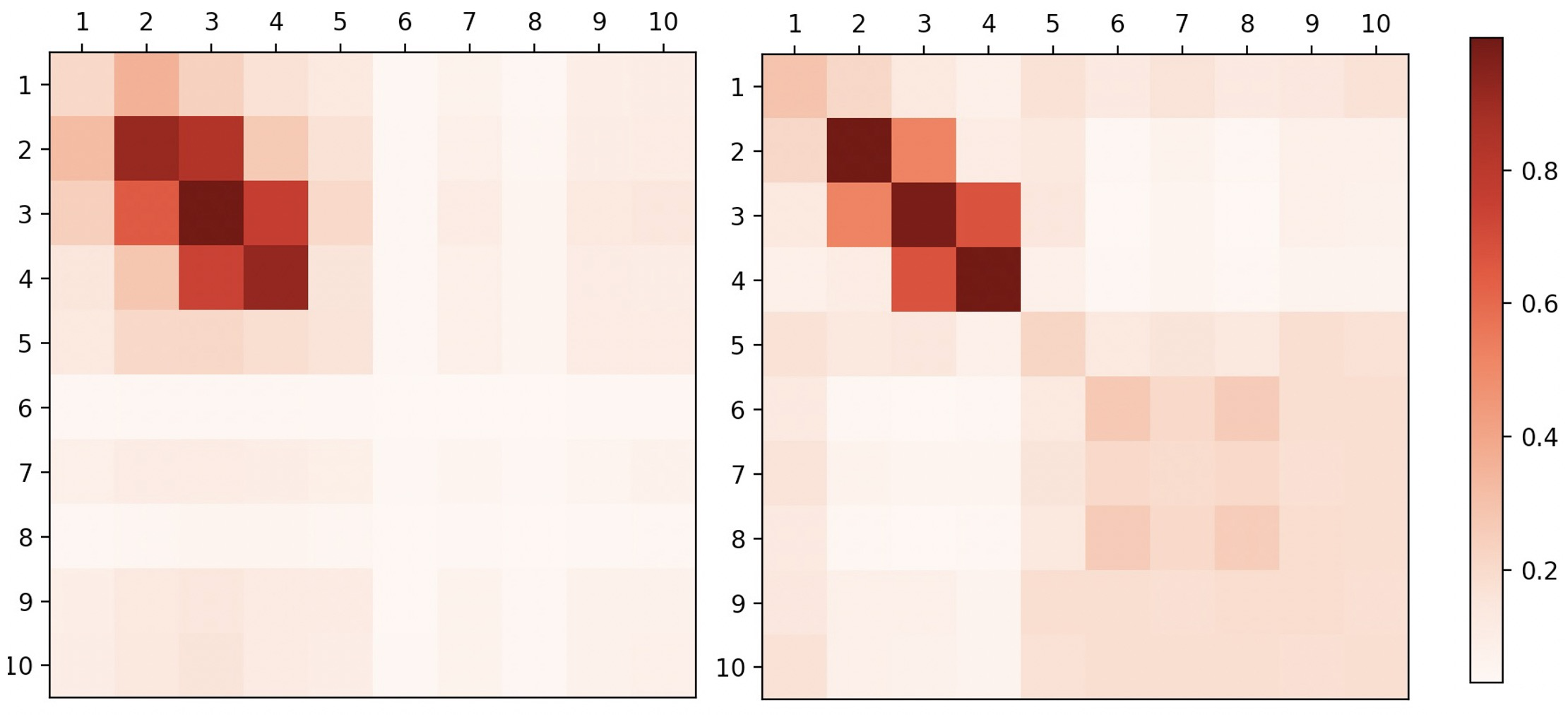}}
\quad
\subfigure[Recovered latent graph across nodes for syn-10 dataset (nodes with weak strength of mutual influence are ignored).]{
    \includegraphics[width=4.5cm]{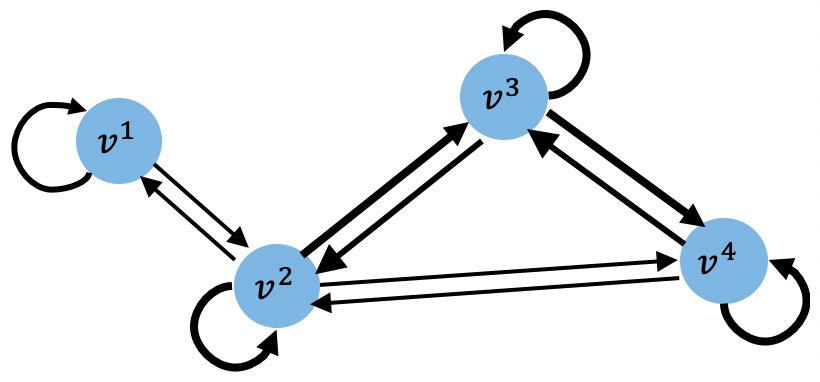}}
\caption{Comparison of infectivity matrix on syn-10d dataset and recovered graph structure.}
\label{fig:causal}
\end{figure*}

\subsubsection{Comparison of Prediction Accuracy on Synthetic and Real Datasets}
As shown in Fig.~\ref{fig:res}, GRPP outperforms the baselines on both synthetic datasets and real datasets. The datasets we adopted vary significantly in the number of nodes, i.e., synthetic-10d has only 10 nodes while Higgs has 900 nodes. Specifically, in all datasets, GRPP improves upon GBTPP by $1.5\%$ - $2\%$ for propagation node prediction and also a notable margin for propagation time prediction. The major innovations that contribute to the advantage of GRPP compared with GBTPP lie in the following: 
\begin{itemize}
    \item GRPP employs a graph propagation mechanism where the node representation preserves the second-order proximity, while GBTPP only preserves the first-order proximity. In short, GRPP captures richer structural properties than GBTPP.
    \item GRPP uses an attentive intensity model to capture long-term dependencies intrinsic to a graph sequence while GBTPP applies a recurrent structure that may fail in modeling interactions between two events located far in the temporal domain. 
\end{itemize}
 
Not surprisingly, neural models (RMTPP, NHP, GBTPP and GRPP) have better performance than the conventional model MHP which makes strong 
assumptions on the generative process of the data while neural models use more expressive recurrent networks to learn the process from the propagation history. We remark that among the neural models, GRPP and GBTPP outperform the others by modeling the structural information of the graph. Similar to the finding in \cite{gbtpp2020}, our result verifies the hypothesis that graph event sequence modeling, as a special case of event sequence modeling, requires a suitable model to incorporate the structural information.

\subsubsection{Model Interpretability - Graph Structure Recovery}

As described in the above section, the node representation contributes to uncover the infectivity matrix, or equivalently the adjacency matrix $\bm{A}$ of $\mathcal{G}$, with the guidance of the graph regularization method. For the illustration purpose, we visualize the infectivity matrix recovered (right) during the learning of \textbf{syn-10d} data, compared with the ground-truth of the generative Hawkes process (left) in Fig.~\ref{fig:causal}. The figure shows that we closely recover the influence between nodes and therefore maintain the model \textbf{interpretability}.

\subsubsection{Ablation Study}
We conduct an ablation study on Meme to validate the effectiveness of key components that contribute to the improved outcomes of our proposed model. We name GRPP removing different components as follows:
\begin{itemize}
    \item \textbf{woGP}: GRPP without graph propagation model or graph regularizer, i.e., we directly pass the sequence encoding vector into the intensity model without any regularization during the training.
    \item \textbf{woAT}: GRPP without latent attentive intensity model. Specially, the excitation and decay factor of the intensity process is directly determined by the closest node embedding instead of a weighted sum of distant node embedding.
\end{itemize}

\begin{figure*}[htbp]
\centering
\subfigure[Propagation node $/$ time prediction.]{
    \includegraphics[width=7cm]{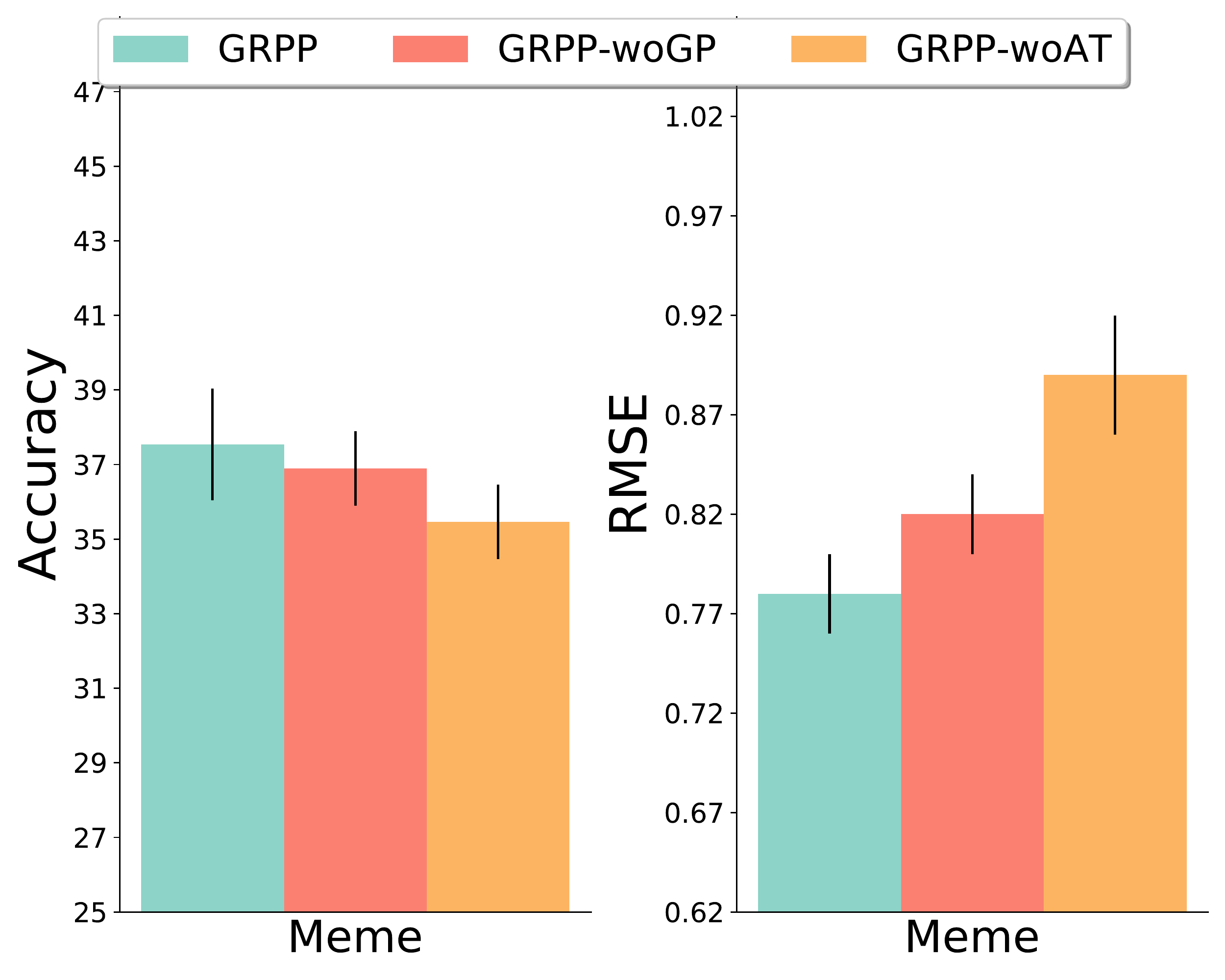}}
\quad
\subfigure[Training curves comparison.]{
    \includegraphics[width=7cm]{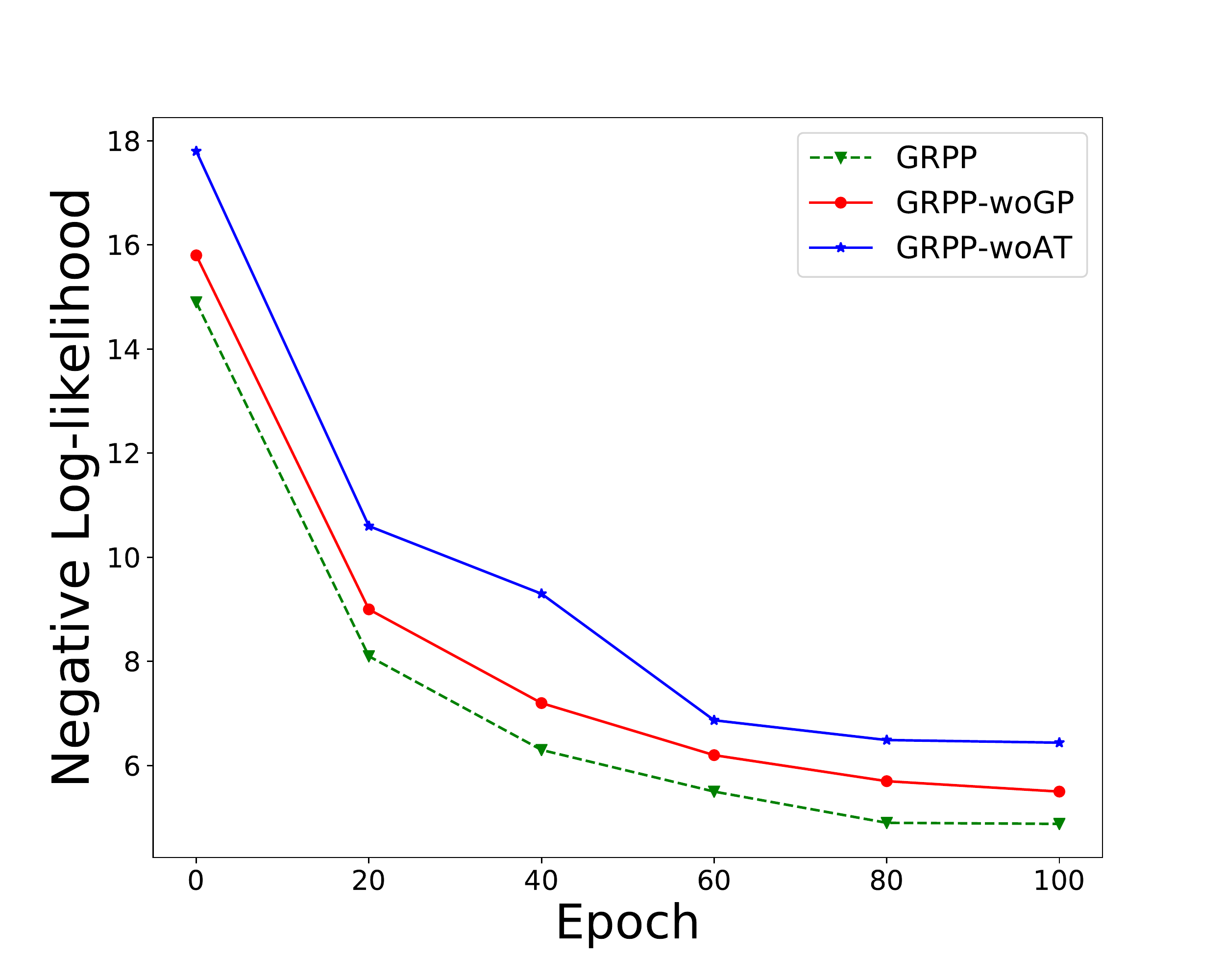}}
\caption{ Predictive performance of variants of GRPP fitted on Meme.}
\label{fig:ablation}
\end{figure*}

Fig.~\ref{fig:ablation} summarizes our experimental results. As shown, we see that both the graph propagation and the latent intensity mechanisms contribute to the model performance. Specifically, GRPP-woAT achieves the worst performance in both training and prediction, which proves the effectiveness of latent intensity model that captures the long-term dependencies. The effect of structural propagation is evident as well: GRPP-woGP achieves worse performance because it fails to incorporate the learnt node representation into the model. Consistent with the discussion above, building the attention over the positional encoder states is not the most suitable model for graph event sequences.


\section{Conclusions}
We presented a novel neural architecture with a graph regularized procedure to model the event propagation process. We employed a graph propagation model and a latent intensity model to capture the structural and temporal dynamics of the event sequence while maintaining the model interpretability. Extensive experiments over real-world datasets have proved the advantages of our model compared to conventional methods and state-of-the-art methods.


\begin{thebibliography}{00}
\bibitem{daley2003}
D. Daley, D. Vera-Jones, An introduction to the theory of point process: general theory and structure, Springer Science \& Business Media, 2003.

\bibitem{du2016recurrent}
N. Du, H. Dai, R. Trivedi, U. Upadhyay, M. Gomez-Rodriguez, L. Song,
``Recurrent marked temporal point processes: embedding event history to vector,''  ACM SIGKDD, pp. 1555--1564, August 2016.

\bibitem{farajtabar2015coevolve}
M. Farajtabar, Y. Wang, M.G. Rodriguez, S. Li,  H. Zha, L. Song, ``Coevolve: A
joint point process model for information diffusion and network co-evolution,'' NIPS, pp. 1954-1962, 2015.

\bibitem{hawkes1971spectra}
A.G. Hawkes, Spectra of some self-exciting and mutually exciting point process, in Biometrika vol. 58(1), pp. 83-90, 1971.


\bibitem{mem2014}
J. Leskovec, ``Snap datasets: Stanford large network dataset collection,'' Stanford, 2014.

\bibitem{graphreg2018}
Liu, Y., Yan, T., Chen, H.: Exploiting graph regularized multi-dimensional hawkes
processes for modeling events with spatio-temporal characteristics. In: IJCAI, pp. 2475–2482 (2018)


\bibitem{mei2017neural}
H. Mei, J.M. Eisner, ``The neural hawkes process: A neurally self-modulating
multivariate point process,'' Advances in neural information processing systems, pp. 6754--6764, 2017.

\bibitem{lecturenote2018}
J.G. Rosmussen, ``Lecture note: temporal point process and the conditional intensity function,'' Aalborg University, pp. 93--102, 2018.

\bibitem{palm1943inten}
C. Palm, Intensitatsschwankungen im fernsprechverker, Ericsson Technics, 1943.

\bibitem{rasmussen2013}
J.G. Rasmussen, ``Bayesian inference for hawkes processes,'' Methodology and
Computing in Applied Probability, vol. 15, pp. 623-642, 2013.

\bibitem{dyrep2019}
R. Trivedi, M. Farajtabar, P. Biswal, H. Zha, ``Dyrep: learning representations
over dynamic graphs,'' International Conference on Learning Representations, 2019.


\bibitem{dyrep2019}
R. Trivedi, M. Farajtabar, P. Biswal, H. Zha, ``Dyrep: learning representations
over dynamic graphs,'' International Conference on Learning Representations, 2019.


\bibitem{li2018}
Y. Li, N. Du, S. Bengio, ``Time-dependent representation for neural sequence prediction,'' International Conference on Learning Representations, 2018.


\bibitem{latent2019}
Q. Wu, Z. Zhang, X. Gao, J. Yan, G. Chen, ``Learning latent process from
high-dimensional event sequences via efficient sampling,'' Advances in Neural Information Processing Systems, pp. 3847--3856, 2019.

\bibitem{gbtpp2020}
W. Wu, H. Liu, X. Zhang, Y. Liu, H. Zha ``Modeling event propagation via
graph biased temporal point process,'' IEEE Transactions on Neural Networks and
Learning Systems, 2020.

\bibitem{rppn2019}
S. Xiao, J. Yan, M. Farajtabar, L. Song, X. Yang, H. Zha, ``Learning time
series associated event sequences with recurrent point process networks,'' IEEE Transactions on Neural Networks And Learning Systems, pp. 3124--3136, 2019.

\bibitem{sapp2020}
Q. Zhang, A. Lipani, O. Kirnap, E. Yilmaz, ``Self-attentive hawkes processes,'' International conference on machine learning, March, 2020.

\bibitem{zhou2013}
K. Zhou, H. Zha, L. Song, ``Learning social infectivity in sparse low-rank networks
using multi-dimensional hawkes processes,'' Artificial Intelligence and Statistics, 2013.

\bibitem{thp2020}
S. Zuo, H. Jiang, Z. Li, T. Zhao, H. Zha, ``Transformer hawkes process,'' International conference on machine learning, March, 2020.


\bibitem{gat2018}
P. Veličkovi\'c, G. Cucurull, A. Casanova, A. Romero, P. Liò, Y. Bengio,
''Graph attention networks,'' International Conference on Learning Representations, 2018.


\bibitem{luong2015}
M. Luong, H. Pham, C. D. Manning. ''Effective Approaches to Attention-based Neural Machine Translation,'' EMNLP, 2015.

\end{thebibliography}
\end{document}